\documentclass[10pt,twocolumn,letterpaper]{article}

\usepackage{cvpr}
\usepackage{times}
\usepackage{epsfig}
\usepackage{graphicx}
\usepackage{amsmath}
\usepackage{amssymb}
\usepackage[numbers,sort&compress]{natbib}
\usepackage{booktabs}
\usepackage{algorithm}
\usepackage{algpseudocode}
\usepackage{amsfonts}
\usepackage{bbm}
\usepackage{tabularx}
\usepackage{booktabs}

\usepackage[breaklinks=true,bookmarks=false]{hyperref}

\cvprfinalcopy 

\def\cvprPaperID{****} 

\begin{document}

\title{Diffusion-Refined Segmentation and Vision-Language Interpretation for Pediatric Brain Tumor MRI}

\renewcommand\thefootnote{\#}
\author{Wentao Ke\thanks{These authors contributed equally.}\\
Department of Mechanical Engineering\\
Stanford University\\
{\tt\small koki8762@stanford.edu}
\and
Jianche Liu\footnotemark[1]\\
School of Medicine\\
Stanford University\\
{\tt\small jianche@stanford.edu}
}

\maketitle
\renewcommand\thefootnote{\arabic{footnote}}

\begin{abstract}
Accurate pediatric brain tumor segmentation remains challenging due to limited annotated data, heterogeneous imaging phenotypes, diffuse tumor boundaries, and class imbalance across tumor subregions. Here, we present a two-stage deep learning framework for improving multimodal pediatric brain MRI segmentation and clinical interpretation. First, we evaluate 3D Res U-Net and Swin-UNETR baselines on BraTS-PEDs MRI scans, using four co-registered modalities to predict tumor core, whole tumor, and enhancing tumor regions. Second, we introduce diffusion-based refinement models conditioned on coarse Swin-UNETR predictions, including a 3D DDPM refiner and MedSegDiff. Conditioning substantially improves diffusion stability and performance, particularly for enhancing tumor boundary segmentation. Conditioned MedSegDiff achieves the strongest boundary agreement with the lowest HD95. Finally, predicted tumor volumes and representative segmentation overlays are integrated with a multimodal language model to generate structured radiology-style reports. Together, our results suggest that coarse-to-refined diffusion segmentation can improve pediatric tumor boundary delineation and support end-to-end interpretable AI-assisted neuro-oncology workflows.

\end{abstract}

\section{Introduction}

\begin{figure}[htb]
    \centering
    \includegraphics[width=1.0\linewidth]{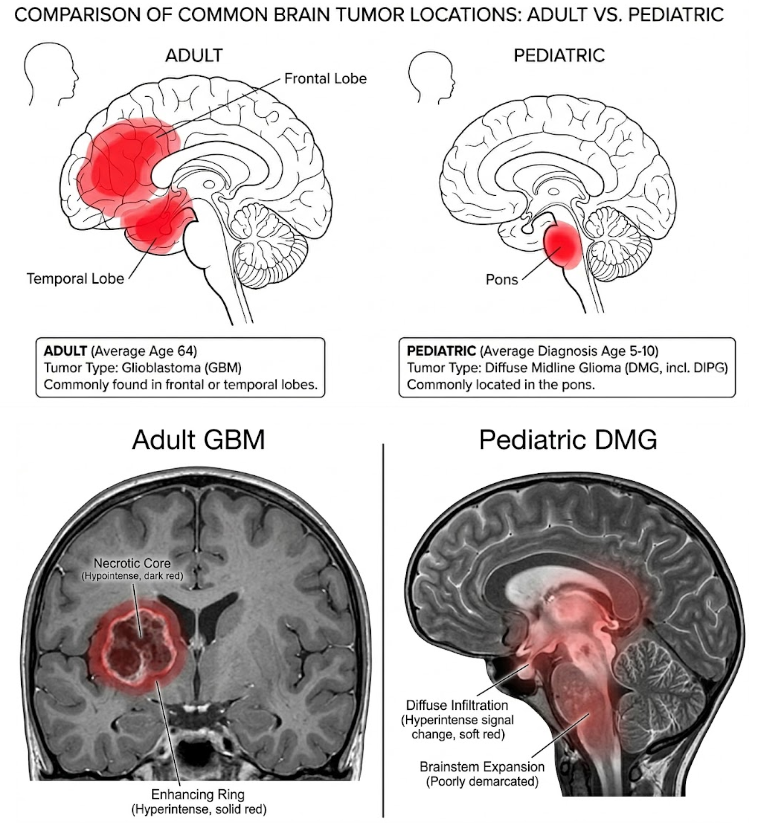}
    \caption{Difference between adult and pediatric brain tumor. Image generated by Gemini 3.1 Pro, based on clinical concepts from \cite{kazerooni2024braintumorsegmentationbrats}.}
    \label{fig:comparison}
\end{figure}

 Accurate delineation of pediatric brain tumor subregions on MRI is essential for diagnosis, treatment planning, response assessment, and longitudinal monitoring. However, comparaed to adult gliomas, pediatric tumors differ substantially in their anatomical distribution, imaging appearance, biology, and heterogeneous tumor subregions, making algorithms trained primarily on adult datasets insufficient for pediatric applications \cite{familiar2024towards}\cite{kazerooni2025brats}. Many pediatric tumors, particularly high-grade gliomas and diffuse midline gliomas, show poorly defined and infiltrative boundaries compared to adult brain tumors ~\ref{fig:comparison}, where edema, enhancing tumor, necrosis, and non-enhancing tumor can overlap or appear subtly on multi-modal MRI images \cite{aboian2017imaging}. This diffuse phenotype increases inter-observer variability and makes manual segmentation time-consuming and difficult to standardize. In addition, pediatric brain MRI datasets are relatively limited because these tumors are rare, and expert voxel-level annotations are costly to obtain. These limitations reduce the robustness and generalizability of conventional supervised deep learning models. The comparison of brain tumor between adults and pediatric and highlight of diffusive nature of pediatric brain tumor is shown in Figure ~\ref{fig:comparison}. 

\begin{figure*}[t]
    \centering
    \includegraphics[width=\textwidth]{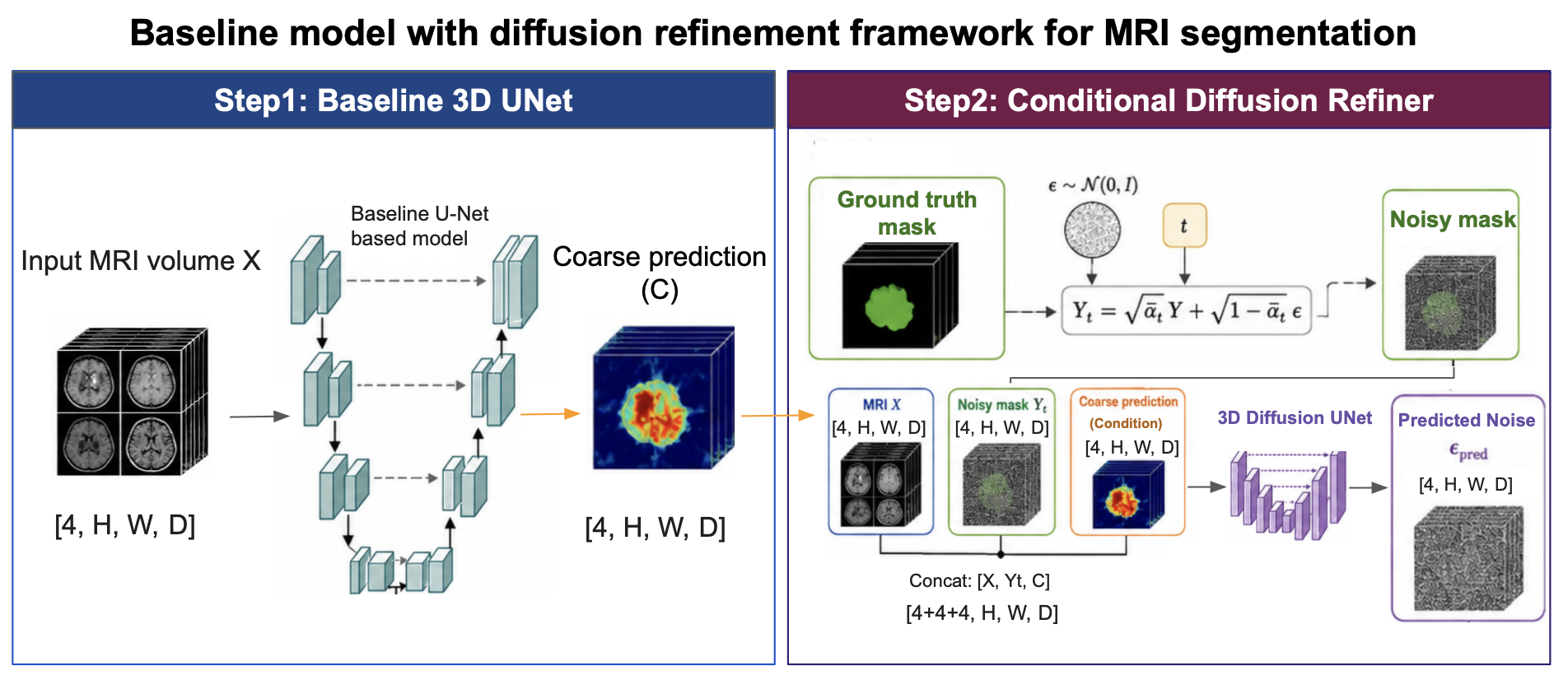}
    \caption{Overview of the model and training setup. We first train a baseline U-Net-based model to generate coarse segmentation predictions. These predictions are then used as conditional inputs for refinement by diffusion-based denoising models.}
    \label{fig:modelsetup}
\end{figure*}

Here, we envision that diffusion-based segmentation refinement offers a promising solution by learning a probabilistic denoising process that can improve coarse tumor masks, sharpen uncertain boundaries, and better model ambiguity in heterogeneous tumor regions. As illustrated in Figure.~\ref{fig:modelsetup}, our framework follows a two-step strategy: first, a baseline U-Net-based segmentation model generates an initial coarse tumor mask; second, this coarse prediction is used as a conditional input to diffusion-based denoising models for mask refinement. Developing such refinement models for pediatric brain tumor segmentation may improve quantitative imaging reliability and support more reproducible clinical and research workflows. In addition, we envision that a state-of-the-art multimodal language foundation model could help interpret the segmentation results and generate human-friendly medical reports.

The input to our algorithm consists of pediatric brain tumor MRI scans; we subsequently employ a hybrid segmentation architecture incorporating a 3D U-Net and a diffusion model for precise tumor delineation, followed by a state-of-the-art multimodal foundation model (e.g., GPT or Gemini) that outputs a comprehensive medical report based on the predicted labels and volumetric data.

\section{Related work}
\subsection{Analysis of BraTS-PEDs 2023}
\paragraph{Manual Delineation vs. Automated AI Workflows}
In assessing the landscape of pediatric brain tumor segmentation, it is essential to first consider whether this task is still performed by hand. Historically, manual delineation by expert neuroradiologists was the clinical gold standard. However, this manual approach is inherently flawed for large-scale analysis. A single 3D MRI volume frequently comprises hundreds of axial slices; manually annotating highly heterogeneous pediatric tumors is incredibly tedious, time-consuming, and highly susceptible to inter-rater variability \cite{familiar2024towards} \cite{kazerooni2025brats}. Consequently, the modern clinical and research paradigm has rapidly shifted toward automated deep-learning workflows to standardize evaluation metrics and reduce the diagnostic workload.
\paragraph{The Current State-of-the-Art (SOTA)}
The current state-of-the-art (SOTA) in automated pediatric tumor segmentation is predominantly defined by the recent BraTS-PEDs 2023 Challenge, which provided the first large-scale benchmarking dataset for this specific domain \cite{kazerooni2025brats}. Based on the challenge results, the SOTA landscape is dominated by self-configuring convolutional frameworks, massive transfer learning, and hybrid Transformer-CNN architectures. Specifically, the top-performing methodology by the CNMCPMI2023 team achieved the highest overall score by employing a sophisticated ensemble of the self-configuring nnU-Net and the transformer-based Swin UNETR, utilizing label-wise ensembling to refine predictions \cite{capellan2024model}. Other SOTA benchmarks were established by automated machine learning (AutoML) frameworks, such as the NVAUTO team utilizing MONAI’s Auto3DSeg to autonomously configure a SegResNet architecture \cite{myronenko2025auto3dseg}.
\paragraph{Strengths and Weaknesses of SOTA Approaches}
An analysis of these SOTA approaches reveals distinct methodological strengths and weaknesses. A major strength of the leading methods is their robustness achieved through heavy ensembling and massive data scaling. For example, the Blackbean team introduced the Scalable and Transferable U-Net (STU-Net), an ultra-large-scale model with 1.4 billion parameters pre-trained on the TotalSegmentator dataset, effectively overcoming the scarcity of pediatric-specific training data \cite{huang2024evaluating}. However, a significant weakness of these approaches is their extreme computational overhead. Massive ensembles (e.g., combining nnU-Net and Swin UNETR) or billion-parameter models like STU-Net require immense GPU memory and inference time, making them highly impractical for real-time, resource-constrained clinical deployment. Furthermore, purely convolutional approaches evaluated in the challenge, while efficient, still struggled with severe class imbalances, particularly in segmenting the extremely small Enhancing Tumor (ET) regions in diffuse midline gliomas.
\paragraph{Current Innovative Methodologies}
Among the participating algorithms, several teams introduced exceptionally clever architectural innovations to solve these specific weaknesses. The SherlockZyb team \cite{zhou2024brain} presented a highly clever extension of the nnU-Net by incorporating self-supervised pre-training coupled with an adaptive region-specific loss function. This approach was highly innovative because it directly addressed the pediatric-specific tumor heterogeneity and class imbalance without needing to rely on external, billion-parameter datasets. Another highly clever and unconventional approach was developed by the UMNIverse team, who introduced the Temporal Cubic PatchGAN (TCuP-GAN). By incorporating Convolutional Long Short-Term Memory networks (ConvLSTMs) into a generative adversarial framework, they creatively treated the multi-parametric MRI sequences as temporal features \cite{mantha2023automated}. This departure from the standard U-Net paradigm demonstrates a highly innovative way to leverage inter-modality relationships for improved segmentation reliability.

\subsection{Diffusion-based Model in Brain Tumor Segmentation}
Diffusion models have recently been explored for brain tumor MRI segmentation as probabilistic refiners. MedSegDiff \cite{wu2023medsegdiff} introduced a DDPM-based medical segmentation framework and showed that iterative denoising can improve brain tumor mask quality, especially by modeling uncertainty and high-frequency boundary structure. MedSegDiff-V2 \cite{wu2024medsegdiffv2} extended this idea with transformer-based conditioning, which explicitly connects noisy mask features with semantic image features rather than simply replacing U-Net with a transformer. Brain-tumor-specific methods such as SF-Diff \cite{mi2025sfdiff}, BTSegDiff \cite{qin2025btsegdiff}, FCFDiff-Net \cite{wu2025fcfdiffnet}, and conditional diffusion segmentation further exploit multimodal MRI guidance, spatial-channel attention, or feature-level diffusion to improve boundary delineation and robustness. Their main strength is that segmentation is treated as a conditional generative process, making them attractive for ambiguous tumor margins and small or discontinuous lesions. 

However, weaknesses for using diffusion remain. First, diffusion inference is slower \cite{song2022denoisingdiffusionimplicitmodels} than CNN/Transformer-based model, especially in the setting where we use 3D dataset. Second, current papers report mainly results on adult BraTS cohorts rather than pediatric tumors, making the baseline missing. Third, the diffusion gains over strong nnU-Net/Swin UNETR ensembles have not yet been consistently established. In current BraTS-style benchmarks, state-of-the-art performance is still largely dominated by well-engineered supervised 3D CNN/Transformer ensembles with careful preprocessing and post-processing.

\section{Methods}
\subsection{Baseline Model: 3D Residual U-Net}
The foundational implementation of our 3D Res U-Net\cite{monai_tutorial_3dunet} was adapted from the official MONAI (Medical Open Network for AI) framework\cite{monai_consortium2022}. This network extends the standard fully convolutional U-Net architecture by integrating residual learning mechanisms into its encoding and decoding pathways, replacing of sequential convolutional layers with residual units. During the forward pass, for any given spatial level l, the residual unit performs an identity mapping via a skip connection to facilitate efficient gradient propagation:
\begin{equation}
    x_{l+1} = ReLU(x_{l} + F(x_{l}, W_{l}))
\end{equation}

To address the extreme spatial class imbalance, where the target enhancing tumor might occupy less than 1\% of the total 3D MRI volume, the network was optimized using the Soft Dice Loss function rather than standard Cross-Entropy. The loss is computed over the predicted probability map $\hat{Y}$ and the ground truth Y: 
\begin{equation}
    \mathcal{L}_{Dice} = 1 - \frac{2 \sum_{c=1}^{K} \sum_{i=1}^{N} Y_{c,i} \hat{Y}_{c,i} + \epsilon}{\sum_{c=1}^{K} \sum_{i=1}^{N} (Y_{c,i}^2 + \hat{Y}_{c,i}^2) + \epsilon}
\end{equation}
where N is the total voxel count, K=4 represents our target anatomical classes, and $\epsilon = 10^{-5}$ is a smoothing factor for numerical stability.

\subsection{Swin-UNETR}
The foundational architecture of Swin-UNETR\cite{monai_tutorial_swinunetr} was sourced from the MONAI framework\cite{monai_consortium2022}, though we developed the multi-modal preprocessing pipeline and the transfer-learning training loop based on 1470 brain diffuse glioma patients. Swin-UNETR reformulates 3D semantic segmentation as a sequence-to-sequence prediction task. The input MRI volume is partitioned into non-overlapping 3D patches and projected into a 1D sequence of embeddings, which act as input tokens for a hierarchical Swin Transformer encoder. This encoder captures profound global contextual information and passes multi-scale features via skip connections to a standard CNN decoder to reconstruct the precise voxel-wise segmentation mask $\hat{Y}$.

To overcome the quadratic computational complexity $(O(N^{2}))$ of standard global attention in massive 3D volumes, Swin-UNETR employs a highly efficient Shifted Window Self-Attention mechanism. Multi-Head Self-Attention (W-MSA) is computed exclusively within local non-overlapping windows. To enable cross-window communication, the subsequent layer shifts the windows by half a window size. The core self-attention for query Q, key K, and value V matrices is formulated as: 
\begin{equation}
    Attention(Q, K, V) = Softmax(\frac{QK^{T}}{d})V\\
\end{equation}
The alternating layer updates are formulated as: 
\begin{gather}
    z^{l} = W\text{-}MSA(LayerNorm(z^{l-1})) + z^{l-1}\\
    z^{l} = MLP(LayerNorm(z^{l})) + z^{l}\\
    z^{l + 1} = SW\text{-}MAS(LayerNorm(z^{l})) + z^{l}
\end{gather}

Furthermore, because Transformers theoretically lack the spatial inductive bias inherent to CNNs, training entirely from scratch on moderately sized datasets is highly prone to overfitting. Therefore, our optimization strategy utilized transfer learning by initializing the Transformer encoder with weights  pre-trained on diverse medical corpora. Our training objective fine-tunes these parameters using the Soft Dice Loss:
\begin{equation}
    \theta^{*}_{fine} = \arg \min E_{(X,Y)}[L_{dice}(F_{\theta}(X), Y)]
\end{equation}
which is initialized at $\theta = \theta_{pre}$. This pre-training initialization drastically accelerates convergence compared to random initialization.

\subsection{U-Net Based Diffusion Refiner}
We cast 3D segmentation as conditional denoising diffusion over the label map
\citep{ho2020ddpm}. Let $I\in\mathbb{R}^{4\times H\times W\times D}$ denote the four
co-registered MRI modalities and $Y_0\in\{0,1\}^{4\times H\times W\times D}$ the one-hot
ground-truth segmentation ($K{=}4$ classes). A forward Markov chain corrupts $Y_0$ with
Gaussian noise,
\begin{equation}
Y_t=\sqrt{\bar\alpha_t}\,Y_0+\sqrt{1-\bar\alpha_t}\,\varepsilon,\qquad
\varepsilon\sim\mathcal{N}(0,\mathbf{I}),
\end{equation}
where $\{\bar\alpha_t\}_{t=1}^{T}$ follow a cosine schedule with $T{=}1000$
\citep{nichol2021improved}. A 3D U-Net $f_\theta$ recovers the clean label (the
$x_0$-parameterization), $\hat Y_0=f_\theta(x_t,t)$, from the noised map stacked with the
conditioning context: $x_t=[\,I\,;\,Y_t\,]$ (8 channels, unconditional) or
$x_t=[\,I\,;\,Y_t\,;\,C\,]$ (12 channels), where $C$ is a 4-class coarse prior generated by a
pretrained SwinU-Net in our baseline model (Swin-UNETR prtrained). \citep{hatamizadeh2022swinunetr}. We minimize a region-aware objective
\begin{equation}
\mathcal{L}=\mathcal{L}_{\mathrm{DiceCE}}(\hat Y_0,y)
+\lambda\sum_{x} w(x)\,\big\lVert\sigma(\hat Y_0)-Y_0\big\rVert^2 ,
\qquad \lambda=0.1,
\end{equation}
coupling a Dice--cross-entropy overlap term with a boundary-weighted MSE ($w(x){=}3$ on class
edges, $1$ elsewhere) that sharpens contours; $\sigma$ is the channel-wise softmax. At
inference the unconditional model samples $Y_T\sim\mathcal{N}(0,\mathbf{I})$ and runs the
reverse chain accelerated by a second-order DPM-Solver \citep{lu2022dpmsolver}. The
conditional model instead injects the prior,
$Y_{t'}=\sqrt{\bar\alpha_{t'}}\,C+\sqrt{1-\bar\alpha_{t'}}\,\varepsilon$, and applies a few
DDIM refinement steps \citep{song2021ddim}, so it refines an existing estimate rather than
generating from scratch, which substantially improves accuracy on the small enhancing-tumor
region.

\subsection{MedSegDiff}
MedSegDiff \citep{wu2023medsegdiff} performs the diffusion on 2D axial slices and corrupts
\emph{only} the segmentation channel, keeping the image as a noise-free condition. With
$I\in\mathbb{R}^{4\times H\times W}$ and one-hot mask $s_0$, the forward process is
\begin{equation}
s_t=\sqrt{\bar\alpha_t}\,s_0+\sqrt{1-\bar\alpha_t}\,\varepsilon,\qquad
\varepsilon\sim\mathcal{N}(0,\mathbf{I}),
\end{equation}
with a linear schedule ($T{=}1000$), and the network input is $x_t=[\,I\,;\,s_t\,]$ (8 ch) or
$x_t=[\,I\,;\,s_t\,;\,C\,]$ (12 ch, conditional). The backbone couples two branches: a diffusion
branch predicts the noise $\hat\varepsilon$, while a parallel \emph{highway} network $G$ (a
lightweight U-Net) maps the image and optional prior to a direct mask $c=G(I,[C])$. Its
multi-scale features are fused into the denoiser through an FF-Parser that filters skip
features in the Fourier domain,
\begin{equation}
\hat F=\mathcal{F}^{-1}\!\big(A\odot\mathcal{F}(F)\big),
\end{equation}
with a learnable spectral attention map $A$ that attenuates high-frequency artifacts. Training
minimizes a noise term plus a calibration term,
\begin{equation}
\mathcal{L}=\mathbb{E}\big\lVert\varepsilon-\hat\varepsilon\big\rVert^2
+\gamma\,\mathcal{L}_{\mathrm{DiceCE}}(c,y),\qquad \gamma=10,
\end{equation}
so the highway learns an accurate segmentation while the diffusion branch models residual
boundary uncertainty. Sampling starts at $s_T\sim\mathcal{N}(0,\mathbf{I})$ and denoises with a
DPM-Solver \citep{lu2022dpmsolver}, fusing the diffusion output with the calibration map.
Conditioning simply appends the pretrained coarse prior $C$, letting the highway refine an
existing prediction rather than segment from scratch, which markedly improves the hardest
enhancing-tumor class and boundary precision.

\subsection{Multimodal Large Language Model (MLLM) Clinical Reporting Pipeline}
While high-fidelity segmentation masks are crucial for quantitative analysis, voxel-wise predictions alone do not directly translate into actionable clinical diagnoses. To bridge the gap between raw algorithmic outputs and clinical utility, we engineered an automated reporting pipeline utilizing MLLMs, such as Gemini 2.5 Pro. 

The interpretation pipeline operates in two sequential stages. First, the continuous probability maps outputted by the Swin-UNETR are thresholded to generate discrete anatomical labels. We then compute the absolute anatomical volume $V_{c}$ for each specific tumor sub-region $c\in {NC, ED, ET}$:
\begin{equation}
    V_{c} = (\sum^{N}_{i = 1} \mathbf{1} (\hat{y_{i}} = C)) \times s_{x} \times s_{y} \times s_{z}
\end{equation}
where $\mathbf{1}$ is the indicator function that equals 1 if voxel i is assigned to class c, and $s_{x}$,$s_{y}$,$s_{z}$ represent the physical voxel spacing dimensions (in millimeters) derived from the MRI metadata.

In the second stage, these extracted quantitative volumetric metrics are structured into a standardized textual prompt. This text, concatenated directly with representative 2D axial MRI slices and their overlaid color-coded segmentation masks, serves as a multimodal input to the MLLM. By simultaneously interpreting the visual spatial distribution of the tumor and its precise geometric measurements, the model synthesizes a human-readable clinical report.

\section{Dataset and Features}
\begin{figure}[htb]
    \centering
    \includegraphics[width=1.0\linewidth]{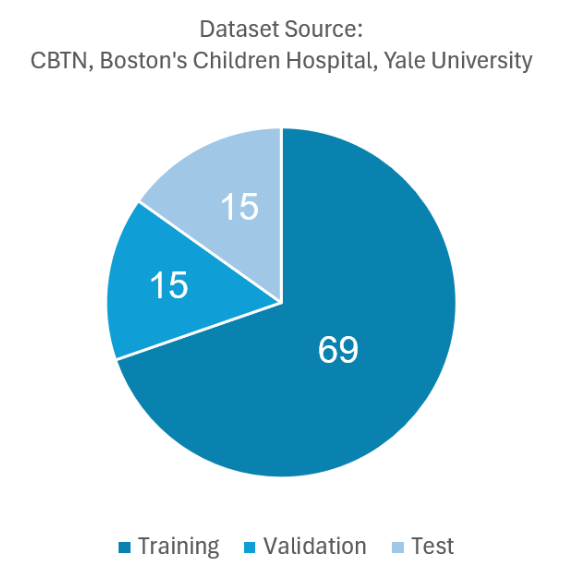}
    \vspace{0.3 cm}
    \includegraphics[width=1.0\linewidth]{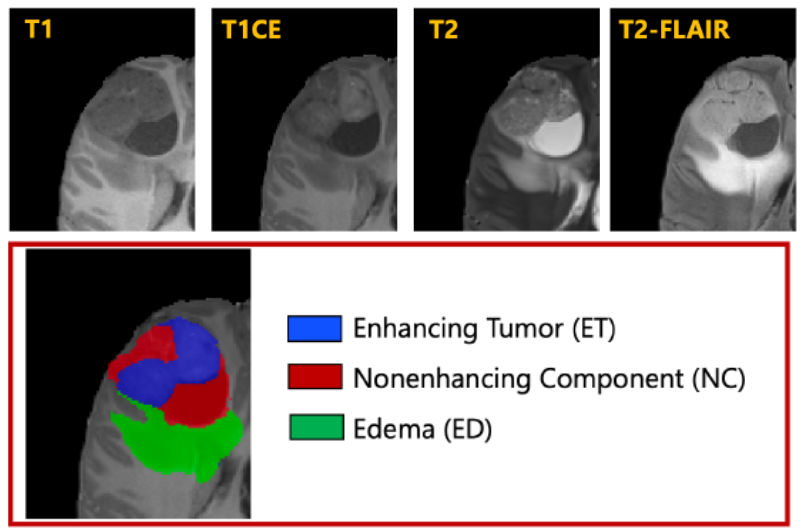}
    \caption{Dataset split, labels and features. Data and image adapted from \cite{kazerooni2024braintumorsegmentationbrats}.}
    \label{fig:datasetup_source}
\end{figure}

The data utilized in this project was sourced from the BraTS-PEDs 2023 Challenge \cite{kazerooni2024braintumorsegmentationbrats}, which provides the first large-scale, multi-institutional cohort of multi-parametric MRI (mpMRI) scans specifically for pediatric high-grade gliomas. On the available pediatric cohort (comprising 99 subjects), we implemented a data split as 69 for training, 15 for validation, and 15 for testing as shown in Figure ~\ref{fig:datasetup_source}.

The input features for each patient consist of four coregistered 3D MRI modalities: native T1-weighted (T1), post-contrast T1-weighted (T1CE), T2-weighted (T2), and T2 Fluid Attenuated Inversion Recovery (T2-FLAIR). The corresponding ground truth annotations provide voxel-wise labels for three mutually exclusive intra-tumoral sub-regions: 
1) \textbf{Enhancing Tumor (ET)}: typically visualized best on T1CE scans; 
2) \textbf{Nonenhancing Component (NC)}: encompassing the necrotic and non-enhancing tumor core; 
3) \textbf{Edema (ED)}: the surrounding water swelling, typically hyperintense on T2-FLAIR scans. 
These distinct regions combined formulate the comprehensive target mask for our segmentation architectures.

Extensive data preprocessing was applied using the MONAI framework. Initially, we utilized the CropForeground operation to crop out the empty, zero-value background air space surrounding the brain, reducing unnecessary computational overhead. Because MRI intensity values are relative and non-standardized across different hospital scanners, we applied NormalizeIntensity(nonzero=True, channel\_wise=True). This step normalizes the voxel intensities to zero mean and unit variance independently for each MRI modality, calculating the statistics strictly on non-zero (brain tissue) voxels to prevent background distortion.

To prevent overfitting on the relatively small training set and to increase the model's robustness to diverse imaging conditions, we implemented 3D geometric and intensity-based data augmentations. Using RandCropByPosNegLabel, we dynamically cropped regions balancing tumor-positive and background-negative samples. Geometrically, we applied RandFlip with a 50\% probability across all three spatial axes (sagittal, coronal, axial), alongside random 90-degree rotations (RandRotate90). Crucially, to simulate the inherent signal strength variability and noise profiles across different clinical MRI scanners, we applied MRI-specific intensity augmentations. Specifically, RandScaleIntensity (with a scaling factor of 0.1) and RandShiftIntensity (with an offset of 0.1) were applied to the input image channels during training. These intensity perturbations ensure that the learned representations are highly invariant to heterogeneous, multi-institutional MRI intensity distributions.

The raw MRI scans inherently possess an isotropic resolution of f$1\times1\times1 mm^{3}$. However, feeding the entire 3D volume into deep Transformer or CNN architectures exceeds standard GPU memory constraints. Consequently, the input resolution during training was stochastically cropped to a fixed spatial size of 96×96×96 voxels. During validation and testing, we reconstructed the full-resolution predictions using a sliding-window inference strategy with a 25\% overlap.

\section{Experiments/Results/Discussion}

\begin{figure}[t]
  \centering
  \includegraphics[width=\linewidth]{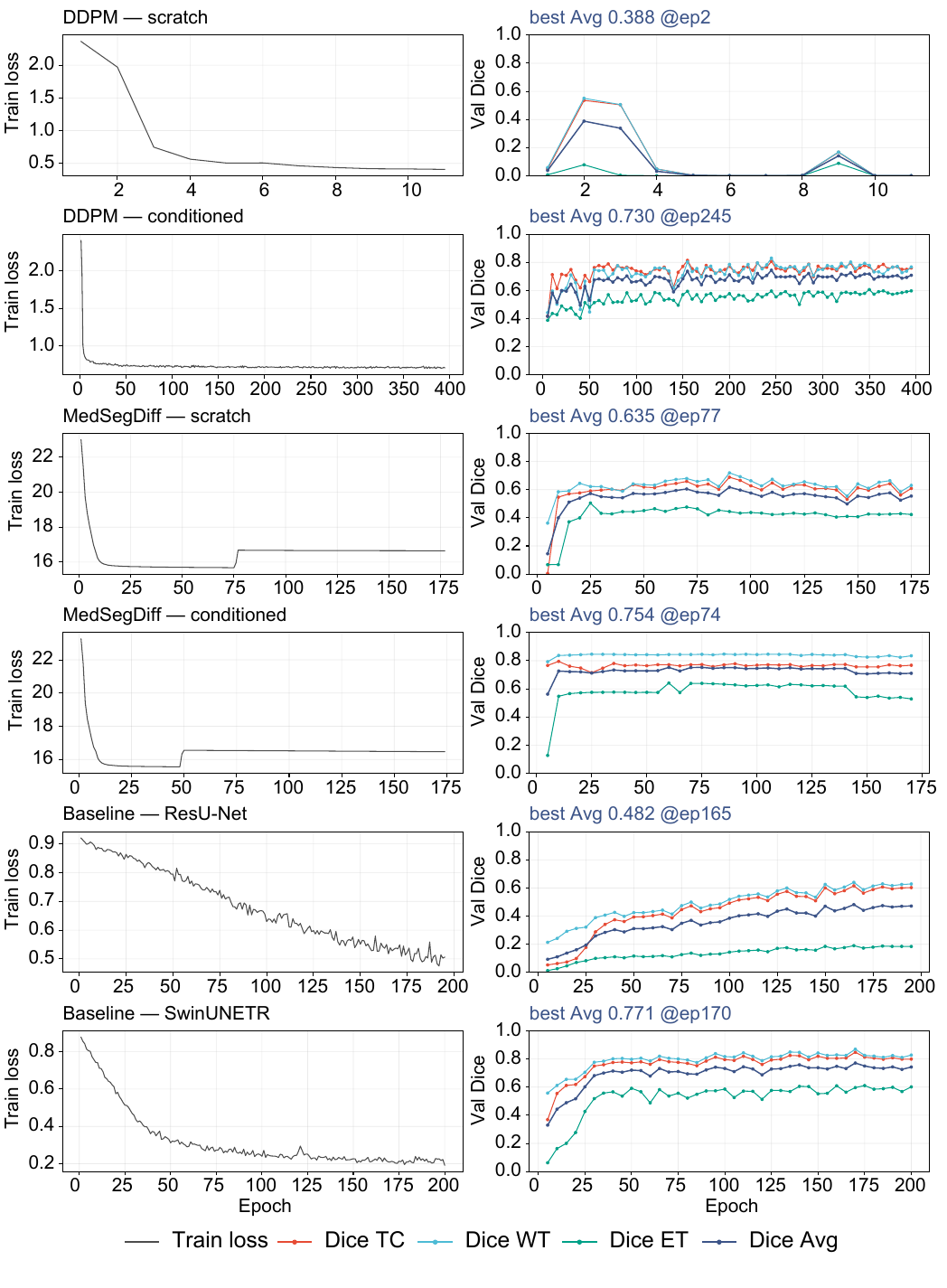}
  \caption{Training loss and validation Dice for the six models.}
  \label{fig:curves}
\end{figure}

\subsection{Training Setup}
\paragraph{3D U-Net Based Models}
Table~\ref{tab:3dunet_hyperparams} summarizes the hyperparameters utilized for training our core baseline and transformer-based segmentation networks. All models were optimized using the AdamW optimizer rather than standard SGD or Adam. We selected AdamW because decoupling the weight decay ($1\times10^{-5}$) from the gradient updates provides significantly better generalization, which is crucial for heavily parameterized architectures like Swin-UNETR. The peak learning rate was empirically set to $1\times10^{-4}$ across all models; this value provided the optimal balance between stable convergence and escaping local minima during the optimization of the highly non-convex Soft Dice Loss landscape.

Due to the extreme computational expense and massive memory footprint of processing 3D multi-modal MRI volumes, batch sizes were strictly dictated by the hardware constraints of a single NVIDIA L4 (24GB VRAM) GPU on Google Cloud Platform (GCP). Specifically, the lightweight 3D Res U-Net accommodated a mini-batch size of 4, whereas the memory-intensive Swin-UNETR was constrained to a batch size of 2. Furthermore, the immense computational cost per epoch precluded the use of traditional k-fold cross-validation. Instead, we utilized a robust single hold-out validation set (15\% of the data) evaluated at regular intervals (e.g., every 5 to 10 epochs). To prevent overfitting, an Early Stopping mechanism was enforced with a patience of 30 epochs.

\begin{table}[t]
\centering
\caption{Training configuration for the core 3D segmentation architectures.}
\label{tab:3dunet_hyperparams}
\small
\renewcommand{\arraystretch}{1.2}
\begin{tabular}{lcc}
\toprule
\textbf{Hyperparameter} & \textbf{3D Res U-Net} & \textbf{Swin-UNETR} \\
\midrule
Spatial domain & $96^3$ patches & $96^3$ patches \\
Optimizer & AdamW & AdamW \\
Peak learning rate & $1 \times 10^{-4}$ & $1 \times 10^{-4}$ \\
Weight decay & $1 \times 10^{-5}$ & $1 \times 10^{-5}$ \\
LR schedule & Constant & Constant / Cosine$^\dagger$ \\
Min. LR (cosine end) & N/A & N/A / $1 \times 10^{-7\dagger}$ \\
Batch size & 4 & 2 \\
Max epochs & 300 & 300 / 150$^\dagger$ \\
Early stopping patience & 30 epochs & 30 epochs / N/A$^\dagger$ \\
Validation interval & 5 epochs & 5 / 10 epochs$^\dagger$ \\
\bottomrule
\end{tabular}
\vspace{2pt}

{\footnotesize Scratch / $^\dagger$ Pre-trained Swin-UNETR.}
\end{table}

\begin{table}[t]
\centering
\caption{Training configuration for the diffusion-based segmentation models.}
\label{tab:training-setup}
\small
\renewcommand{\arraystretch}{1.15}
\begin{tabular}{lcc}
\toprule
\textbf{Hyperparameter} & \textbf{U-Net DDPM} & \textbf{MedSegDiff} \\
\midrule
Spatial domain & $96^3$ patches (3D) & $128^2$ slices (2D) \\
Optimizer & AdamW & AdamW \\
Peak learning rate & $1.00 \times 10^{-4}$ & $4.00 \times 10^{-4}$ \\
Weight decay & $1.00 \times 10^{-5}$ & $0.00$ \\
LR schedule & warmup $+$ cosine & warmup $+$ cosine \\
Warmup epochs & $10$ & $10$ \\
Minimum LR & $1.00 \times 10^{-6}$ & $1.00 \times 10^{-6}$ \\
Batch size & $2$ & $32$ \\
Maximum epochs & $1000$ / $300$\textsuperscript{$\dagger$} & $300$ \\
Tolerance epochs & $150$ & $100$ \\
Gradient clip & $1.00$ max-norm & $1.00$ max-norm \\
Diffusion steps $T$ & $1000$ cosine & $1000$ linear \\
Validation interval & every $1$ epoch & every $1$ epoch \\
\bottomrule
\end{tabular}
\vspace{2pt}

{\footnotesize \textsuperscript{$\dagger$}Conditioned / scratch DDPM.}
\end{table}

\paragraph{Diffusion Models} All four diffusion models were trained with the AdamW optimizer and a
\emph{linear-warmup--cosine-annealing} learning-rate schedule: the rate increases
linearly from $10^{-6}$ to its peak over the first $10$ epochs, then decays following a
cosine curve to $10^{-6}$ over the remaining budget. Gradients are clipped to a maximum
$\ell_2$ norm of $1.0$, validation is run every epoch, and the checkpoint with the highest
mean validation Dice is retained. Training stops early when the mean Dice fails to improve
for a fixed patience (the \emph{tolerance epochs}), and all runs use the same
$69/15/15$ subject split for train/validation/test.

The two model families differ in spatial granularity, which dictates their batch sizes and
peak learning rates. The U-Net DDPM operates on $96^3$ 3D patches (batch size $2$, peak
LR $1\times10^{-4}$, weight decay $10^{-5}$) over a cosine noise schedule with $T{=}1000$
timesteps. MedSegDiff operates on $128\times128$ 2D axial slices, which permits a much larger
batch ($32$) and a higher peak LR ($4\times10^{-4}$, no weight decay) over a linear noise
schedule, also with $T{=}1000$. The unconditional (\emph{scratch}) and coarse-conditioned
variants of each family share identical optimization settings; conditioning only augments the
input with the four coarse-prediction channels. Table~\ref{tab:training-setup} summarizes the
full configuration for diffusion-based models.

\paragraph{MLLM Clinical Report Generation}
For the clinical reporting pipeline, MLLM inference was performed using the Gemini 2.5 Pro foundation model via the Google Cloud Vertex AI API. To ensure high factual accuracy and minimize medical hallucination, generation parameters were kept to their deterministic defaults, utilizing a zero-shot multimodal prompting strategy composed of both the 2D axial MRI slices and the calculated sub-region volumes.

\begin{table*}[h!]
\centering
\caption{Quantitative segmentation results on the pediatric brain tumor test set. Best results for each metric are shown in bold.}
\label{tab:segmentation-results}
\small
\begin{tabular}{lccccccc}
\toprule
Model & Dice TC & Dice WT & Dice ET & Dice Avg & HD95 $\downarrow$ & Sensitivity & Precision \\
\midrule
3D Res U-Net
& 0.62 & 0.64 & 0.19 & 0.48 & 38.35 & \textbf{0.93} & 0.34 \\

Swin-UNETR
& \textbf{0.85} & \textbf{0.87} & 0.60 & 0.77 & 13.02 & 0.84 & \textbf{0.88} \\

Swin-UNETR(from pretrain)
& 0.83 & 0.86 & \textbf{0.63} & \textbf{0.77} & 9.85 & 0.89 & 0.85 \\

DDPM (scratch)
& 0.54 & 0.55 & 0.08 & 0.39 & 102.76 & 0.49 & 0.37 \\

DDPM (conditioned)
& 0.78 & 0.78 & 0.63 & 0.73 & 33.54 & 0.87 & 0.74 \\

MedSegDiff (scratch)
& 0.68 & 0.71 & 0.48 & 0.62 & 49.67 & 0.77 & 0.67 \\

MedSegDiff (conditioned)
& 0.78 & 0.84 & 0.61 & 0.75 & \textbf{9.35} & 0.83 & 0.80 \\
\bottomrule
\end{tabular}
\vspace{2pt}
\\[-2pt]
{\footnotesize TC: tumor core; WT: whole tumor; ET: enhancing tumor. Lower HD95 indicates better boundary agreement.}
\end{table*}

\subsection{Evaluation Metrics}
\label{sec:eval-metrics}

Following the BraTS evaluation protocol, predictions over the four classes
(background, necrotic core, edema, enhancing tumor) are first merged into three
nested binary sub-regions: the \emph{whole tumor} (WT, all tumor classes), the
\emph{tumor core} (TC, necrotic core $\cup$ enhancing tumor), and the
\emph{enhancing tumor} (ET). For each sub-region $r\in\{\mathrm{TC},\mathrm{WT},\mathrm{ET}\}$
we compare the predicted binary mask $P_r$ with the ground-truth mask $G_r$ and report four
complementary metrics: the Dice similarity coefficient (region overlap), the $95$th-percentile
Hausdorff distance (boundary agreement), and sensitivity and precision (voxel-wise detection).
Let $\mathrm{TP}=|P_r\cap G_r|$, $\mathrm{FP}=|P_r\setminus G_r|$, and
$\mathrm{FN}=|G_r\setminus P_r|$ denote the true-positive, false-positive, and false-negative
voxel counts. All metrics are computed per volume and averaged over subjects.

\paragraph{Dice Similarity Coefficient (DSC).}
The Dice score measures the overlap between prediction and ground truth,
\begin{equation}
\mathrm{DSC}(P_r,G_r)=\frac{2\,|P_r\cap G_r|}{|P_r|+|G_r|}
=\frac{2\,\mathrm{TP}}{2\,\mathrm{TP}+\mathrm{FP}+\mathrm{FN}}\in[0,1].
\end{equation}
We report Dice for each sub-region ($\mathrm{Dice}_{\mathrm{TC}}$, $\mathrm{Dice}_{\mathrm{WT}}$,
$\mathrm{Dice}_{\mathrm{ET}}$) together with their mean,
\begin{equation}
\mathrm{Dice}_{\mathrm{Avg}}=\tfrac{1}{3}\big(\mathrm{Dice}_{\mathrm{TC}}
+\mathrm{Dice}_{\mathrm{WT}}+\mathrm{Dice}_{\mathrm{ET}}\big).
\end{equation}

\paragraph{Sensitivity and Precision.}
Sensitivity (recall) is the fraction of true tumor voxels recovered, and precision is the
fraction of predicted tumor voxels that are correct:
\begin{equation}
\mathrm{Sensitivity}=\frac{\mathrm{TP}}{\mathrm{TP}+\mathrm{FN}},\qquad
\mathrm{Precision}=\frac{\mathrm{TP}}{\mathrm{TP}+\mathrm{FP}}.
\end{equation}
Together they expose the over-/under-segmentation behaviour that Dice alone can mask.

\paragraph{95th-percentile Hausdorff distance (HD95).}
The Hausdorff distance measures the worst-case discrepancy between the prediction and
ground-truth surfaces $\partial P_r$ and $\partial G_r$. To reduce sensitivity to outliers,
we use the $95$th percentile of the bidirectional surface distances,
\begin{equation}
\begin{split}
\mathrm{HD95}(P_r,G_r)=\max\Big(
&\underset{p\in\partial P_r}{\mathrm{P}_{95}}\min_{g\in\partial G_r}\lVert p-g\rVert,\\
&\underset{g\in\partial G_r}{\mathrm{P}_{95}}\min_{p\in\partial P_r}\lVert g-p\rVert\Big)
\end{split}
\end{equation}

where $\mathrm{P}_{95}$ denotes the $95$th percentile and $\lVert\cdot\rVert$ the Euclidean
distance (in voxels). Lower HD95 indicates tighter boundary agreement.

\subsection{Results \& Discussion}

\begin{figure}[t]
  \centering
  \includegraphics[width=\linewidth]{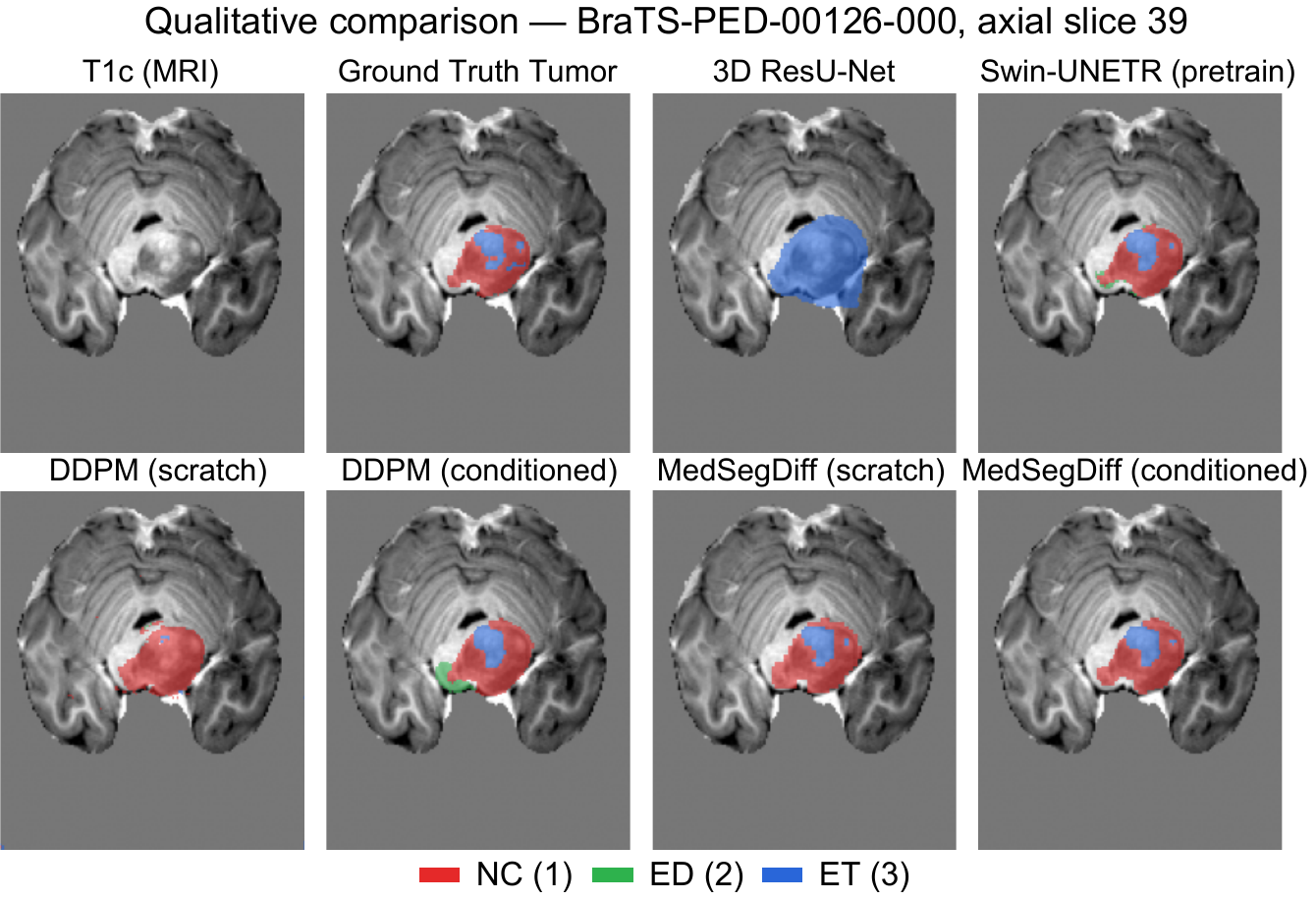}
  \caption{Qualitative comparison on a representative case
  (BraTS-PED-00126, axial slice with the largest tumor extent), shown over the
  T1c image. Overlays denote the necrotic core (NC), edema (ED), and enhancing
  tumor (ET). Conditioned models track the ground truth most closely, whereas
  the unconditional DDPM over-segments.}
  \label{fig:qualitative}
\end{figure}

\paragraph{Conditioning on Predicted Prior is Important for Diffusion Performance}

Table~\ref{tab:segmentation-results} reports quantitative performance, and
Figure~\ref{fig:curves} the corresponding training dynamics, nd Figure~\ref{fig:qualitative}
a representative qualitative comparison. The clearest trend
is that \emph{coarse conditioning is decisive for the performance of diffusion models}. Without a prior, the
unconditional variants are weak and unstable: DDPM (scratch) reaches only $0.39$ average Dice
(enhancing tumor, ET, $0.08$), and MedSegDiff (scratch) $0.62$. Conditioning on the
pretrained Swin-UNETR coarse prediction lifts these to $0.73$ and $0.75$ average Dice
respectively, with ET rising from $0.08$ to $0.63$ for DDPM. Figure~\ref{fig:curves}
explains the reason: the unconditional runs are unstable---DDPM (scratch) collapses after a brief
early peak, and MedSegDiff (scratch) plateaus lower---whereas the conditioned models converge
faster and more stably. This is mirrored
in Figure~\ref{fig:qualitative}, where the unconditional DDPM paints a noisy, over-extended
mask, while the conditioned model recovers a coherent tumor that follows the underlying
anatomy. The coarse prior thus supplies structure the reverse process struggles to learn
from noise alone, especially for the small ET region. 

\paragraph{Boundary Refinement Improvement from Diffusion Model}
As shown in Table~\ref{tab:segmentation-results}, we found that diffusion's strength is \emph{boundary refinement}. MedSegDiff (conditioned)
attains the best HD95 of any model ($9.35$), \emph{lower} than the Swin-UNETR (pretrained) prior
it refines ($9.85$) and far below its scratch counterpart ($49.67$), while retaining competitive
Dice ($0.75$) and balanced sensitivity/precision ($0.83/0.80$). This indicates that the reverse
process tightens contours \emph{beyond} the prior rather than merely reproducing it. In contrast,
DDPM (conditioned) achieves strong Dice (ET $0.63$) but a high HD95 ($33.54$), revealing residual
boundary noise and clear headroom for further refinement through additional reverse steps or
stronger priors.

Finally, we contextualize against the BraTS-PEDs 2023 challenge~\citep{kazerooni2025brats},
whose top teams reported WT Dice $0.81$--$0.84$, TC $0.77$--$0.81$, and ET $0.53$--$0.65$. Our best
region-wise results (Swin-UNETR-pretrained, $0.77$ average; MedSegDiff-conditioned, WT $0.84$ /
ET $0.61$, HD95 $9.35$) fall within this range. We note the comparison is indicative rather than
head-to-head: the challenge uses lesion-wise metrics on its held-out test set, whereas we report
region-wise metrics on an internal split. Nonetheless, our single conditioned diffusion model
matches strong CNN baselines on boundary accuracy without the large nnU-Net/Swin-UNETR ensembles
used by challenge winners.

\paragraph{Qualitative Clinical Evaluation and MLLM Reasoning}

Beyond voxel-wise segmentation accuracy, our pipeline translated raw geometric predictions into actionable clinical insights. We utilized the Gemini 2.5 Pro foundation model via Google Vertex AI to interpret the precise volumetric metrics (extracted from Swin-UNETR) alongside the 2D axial MRI morphology.

Figure ~\ref{fig:gemini_report} demostrate a conversation between the user and Gemini 2.5 Pro to generate an unedited, end-to-end clinical radiology report for patient BraTS-PED-00119-000:

\begin{figure}[htb]
    \centering
    \includegraphics[width=1.0\linewidth]{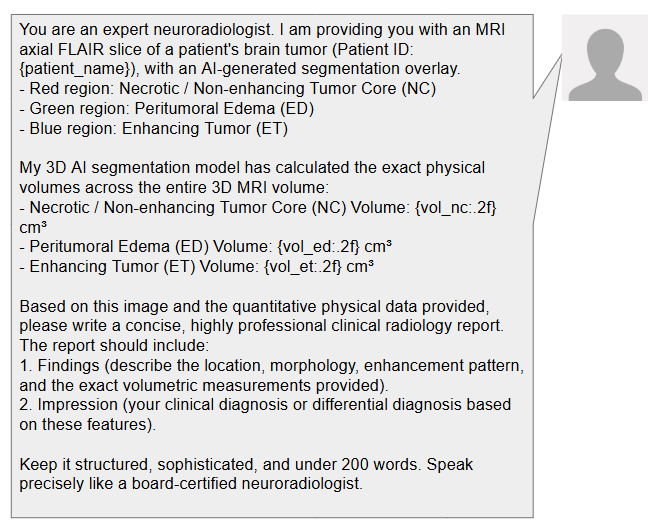}
    
    \vspace{0.3cm} 
    
    \includegraphics[width=1.0\linewidth]{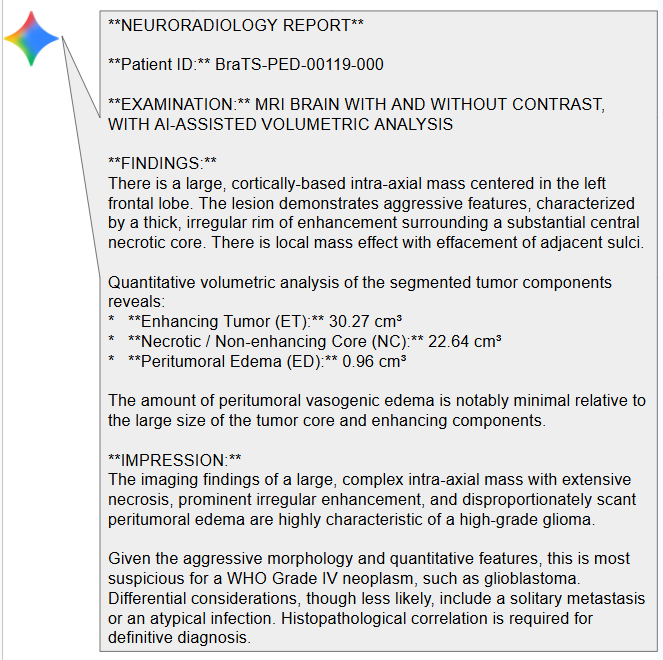}
    
    \caption{End-to-end AI-assisted clinical radiology report generation. (Top) The structured multimodal user prompt combining volumetric metrics and instructions. (Bottom) The final diagnostic report generated by Gemini 2.5 Pro.}
    \label{fig:gemini_report}
\end{figure}

By cross-referencing the immense structural volume of the necrotic core (NC: 22.64 cm³) with the disproportionately minimal peritumoral edema (ED: 0.96 cm³), the model accurately deduced the aggressive morphology of the lesion. Consequently, it formulated a highly professional impression, generating a plausible diagnostic hypothesis that suspects a WHO Grade IV glioblastoma, alongside clinically sound differential considerations (such as solitary metastasis). While definitive diagnosis inherently requires histopathological validation, this case highlights the transformative potential of combining precise 3D topological segmentation with the zero-shot reasoning capabilities of foundation models to assist and accelerate clinical diagnostic workflows.

\section{Conclusion \& Future Work}
\subsection{Conclusion}
In this project, we developed a comprehensive deep learning pipeline for pediatric brain tumor segmentation, systematically evaluating architectures ranging from traditional 3D CNNs to Transformers and generative Diffusion Models. Our quantitative results revealed a clear performance hierarchy. The baseline 3D Res U-Net struggled significantly with complex topologies (Dice Avg: 0.48). Conversely, the transformer-based Swin-UNETR established itself as the most robust segmentation engine, achieving the highest voxel-wise accuracy (Dice Avg: 0.77, Dice WT: 0.87) by leveraging shifted-window self-attention to capture long-range global anatomical context. Furthermore, our exploration into generative AI revealed that while unconditioned DDPMs failed to generate meaningful masks, conditionally guided Diffusion Models (e.g., MedSegDiff) excelled in boundary delineation, achieving the lowest Hausdorff Distance (HD95: 9.35) by iteratively refining the morphological margins of the tumor components. Building upon these robust visual features, we integrated a Multimodal Large Language Model (Gemini 2.5 Pro) as a downstream clinical tool. By jointly interpreting the predicted geometric volumes and 2D MRI morphology, the MLLM successfully bridged the gap between raw pixel-level predictions and actionable clinical insights, generating evidence-based plausible diagnostic hypotheses (e.g., WHO Grade IV glioblastoma).
\subsection{Limitations}
Several critical limitations were observed during our training workflow. First, the severe class imbalance inherent to pediatric gliomas remains a profound challenge; the Enhancing Tumor (ET) core occupies a minuscule fraction of the overall brain volume, making it notoriously difficult to segment even for advanced architectures. Second, 3D Transformer models impose a massive computational and memory footprint, rigidly restricting our training batch size to 2 on a 24GB L4 GPU, which inherently limits gradient stability. Lastly, while diffusion models yielded the sharpest boundaries, their iterative reverse Markov process requires hundreds of sampling steps, making their inference speed prohibitively slow for real-time clinical applications. On the MLLM front, the clinical reporting pipeline is inherently vulnerable to error propagation; any inaccuracies in the upstream segmentation volumes are directly magnified in the MLLM's diagnostic reasoning. Furthermore, relying entirely on the zero-shot capabilities of general-purpose foundation models carries a fundamental risk of medical hallucination, as the generated reports lack strictly verified clinical grounding and could mislead non-expert users without rigorous histopathological correlation.
\subsection{Future Works}
Addressing these computational and morphological bottlenecks will be the primary focus of future work. To mitigate the slow inference of diffusion networks, advanced acceleration techniques such as Denoising Diffusion Implicit Models (DDIM)\cite{song2022denoisingdiffusionimplicitmodels} or Consistency Trajectory Models could be implemented. To combat class imbalance and hardware constraints, exploring patch-based self-supervised pre-training or memory-efficient linear attention mechanisms could further boost the ET Dice score. Finally, to mitigate the hallucination risks of the MLLM, future pipelines should incorporate Retrieval-Augmented Generation (RAG) \cite{lewis2021retrievalaugmentedgenerationknowledgeintensivenlp}, grounding the foundation model's outputs with verified medical literature and incorporating explicit human-in-the-loop (HITL) clinician verification mechanisms before finalizing any radiology report.

\section{Appendix}
\paragraph{Contribution Statement}
Both members contributed equally. Specific responsibilities were divided as follows:

\textbf{Wentao Ke:}

- Engineered the 3D Res U-Net baseline.

- Adapted and optimized Swin-UNETR architectures.

- Developed the MLLM radiology reporting pipeline.

\textbf{Jianche Liu:}

- Implemented unconditional and conditional DDPM.

- Integrated the MedSegDiff generative architecture.

- Optimized diffusion reverse-sampling processes.

\textbf{Shared contributions:}

- Formulated project scope and objectives.

- Conducted literature review.

- Preprocessed multi-modal pediatric dataset.

- Prepared milestone slides and final report.

\paragraph{Acknowledgements}
We would like to express our sincere gratitude to the entire CS231n teaching staff for their invaluable support throughout this quarter. We specifically thank the course administration for generously providing Google Cloud Platform (GCP) and Amazon Web Services (AWS) cloud computing credits. Training the massive 3D Transformer and Diffusion models utilized in this study would not have been feasible without access to these high-performance computational resources.

Additionally, we extend a very special thank you to our Course Assistants (CA), Wenlong Huang and Bailey Trang. We deeply appreciate your continuous guidance, patient listening, and insightful technical feedback during our milestone check-ins, which significantly helped us overcome various implementation hurdles and steered the project in the right direction.

\paragraph{GenAI Usage Statement}

Generative AI tools, including ChatGPT and Gemini, were utilized extensively throughout various stages of this project. Specifically, these foundation models assisted with:

- Implementation: Debugging PyTorch tensor operations and troubleshooting GPU memory constraints during training.

- Writing \& Formatting: Proofreading the final report for grammatical clarity and assisting with complex LaTeX formatting (e.g., table structures and multi-column figure placements).

- Brainstorming: Aiding in the conceptualization of the Multimodal LLM integration pipeline.

Furthermore, aside from acting as developmental assistants, Gemini 2.5 Pro was explicitly integrated as a core experimental component within our proposed clinical pipeline to synthesize automated diagnostic radiology reports.

{\small
\bibliographystyle{ieee}
\bibliography{egbib}

@article{kazerooni2025brats,
   title={BraTS-PEDs: Results of the Multi-Consortium International Pediatric Brain Tumor Segmentation Challenge 2023},
   author={Fathi Kazerooni, Anahita and Khalili, Nastaran and Liu, Xinyang and Haldar, Debanjan and others},
   journal={Machine Learning for Biomedical Imaging},
   volume={3},
   pages={72--87},
   year={2025},
   doi={10.59275/j.melba.2025-f6fg}
}

@article{familiar2024towards,
  title={Towards consistency in pediatric brain tumor measurements: Challenges, solutions, and the role of artificial intelligence-based segmentation},
  author={Familiar, Ariana M and Fathi Kazerooni, Anahita and Vossough, Arastoo and Ware, Jeffrey B and others},
  journal={Neuro-Oncology},
  volume={26},
  number={9},
  pages={1557--1571},
  year={2024},
  doi={10.1093/neuonc/noae093}
}

@inproceedings{capellan2024model,
  title={Model Ensemble for Brain Tumor Segmentation in Magnetic Resonance Imaging},
  author={Capell{\'a}n-Mart{\'\i}n, D and Jiang, Z and Parida, A and Liu, X and others},
  booktitle={Brain Tumor Segmentation, and Cross-Modality Domain Adaptation for Medical Image Segmentation (BraTS 2023)},
  pages={221--232},
  year={2024},
  publisher={Springer}
}

@article{myronenko2025auto3dseg,
  title={Auto3DSeg for Brain Tumor Segmentation from 3D MRI in BraTS 2023 Challenge}, 
  author={Myronenko, Andriy and Yang, Dong and He, Yufan and Xu, Daguang},
  journal={arXiv preprint arXiv:2510.25058},
  year={2025}
}

@inproceedings{huang2024evaluating,
  title={Evaluating STU-Net for Brain Tumor Segmentation},
  author={Huang, Z and others},
  booktitle={Brain Tumor Segmentation, and Cross-Modality Domain Adaptation for Medical Image Segmentation (BraTS 2023)},
  year={2024},
  publisher={Springer}
}

@inproceedings{zhou2024brain,
  title={Brain Tumor Segmentation Based on Self-supervised Pre-training and Adaptive Region-Specific Loss},
  author={Zhou, Y and Zhong, L and Wang, G},
  booktitle={Brain Tumor Segmentation, and Cross-Modality Domain Adaptation for Medical Image Segmentation (BraTS 2023)},
  year={2024},
  publisher={Springer}
}

@article{mantha2023automated,
  title={Automated 3D Tumor Segmentation using Temporal Cubic PatchGAN (TCuP-GAN)}, 
  author={Mantha, Kameswara Bharadwaj and Sankar, Ramanakumar and Fortson, Lucy},
  journal={arXiv preprint arXiv:2311.14148},
  year={2023}
}

@misc{monai_tutorial_3dunet,
  author = {{Project MONAI}},
  title = {3D U-Net Segmentation Tutorial in MONAI},
  year = {2024},
  publisher = {GitHub},
  journal = {GitHub repository},
  howpublished = {\url{https://github.com/Project-MONAI/tutorials/blob/main/3d_segmentation/unet_segmentation_3d_ignite.ipynb}},
  note = {Accessed: 2024-06-05}
}

@article{monai_consortium2022,
  title={MONAI: An open-source framework for deep learning in healthcare},
  author={Cardoso, M Jorge and Li, Wenqi and Brown, Richard and Ma, Nic and Kerfoot, Eric and Wang, Yiheng and Murrey, Benjamin and Myronenko, Andriy and Zhao, Can and Yang, Dong and others},
  journal={arXiv preprint arXiv:2211.02701},
  year={2022}
}

@misc{monai_tutorial_swinunetr,
  author = {{Project MONAI}},
  title = {Swin UNETR BraTS 21 Segmentation Tutorial in MONAI},
  year = {2024},
  publisher = {GitHub},
  journal = {GitHub repository},
  howpublished = {\url{https://github.com/Project-MONAI/tutorials/blob/main/3d_segmentation/swin_unetr_brats21_segmentation_3d.ipynb}},
  note = {Accessed: 2024-06-05}
}

@article{kazerooni2024braintumorsegmentationbrats,
  title={The Brain Tumor Segmentation (BraTS) Challenge 2023: Focus on Pediatrics (CBTN-CONNECT-DIPGR-ASNR-MICCAI BraTS-PEDs)}, 
  author={Fathi Kazerooni, Anahita and Khalili, Nastaran and Liu, Xinyang and Haldar, Debanjan and Jiang, Zhifan and Anwar, Syed Muhammed and others},
  journal={arXiv preprint arXiv:2305.17033},
  year={2024}
}

@article{song2022denoisingdiffusionimplicitmodels,
  title={Denoising Diffusion Implicit Models},
  author={Song, Jiaming and Meng, Chenlin and Ermon, Stefano},
  journal={arXiv preprint arXiv:2010.02502},
  year={2022}
}

@article{lewis2021retrievalaugmentedgenerationknowledgeintensivenlp,
  title={Retrieval-Augmented Generation for Knowledge-Intensive NLP Tasks},
  author={Lewis, Patrick and Perez, Ethan and Piktus, Aleksandra and others},
  journal={arXiv preprint arXiv:2005.11401},
  year={2021}
}

@inproceedings{ho2020ddpm,
  title     = {Denoising Diffusion Probabilistic Models},
  author    = {Ho, Jonathan and Jain, Ajay and Abbeel, Pieter},
  booktitle = {Advances in Neural Information Processing Systems (NeurIPS)},
  year      = {2020}
}

@inproceedings{nichol2021improved,
  title     = {Improved Denoising Diffusion Probabilistic Models},
  author    = {Nichol, Alexander Quinn and Dhariwal, Prafulla},
  booktitle = {International Conference on Machine Learning (ICML)},
  year      = {2021}
}

@inproceedings{song2021ddim,
  title     = {Denoising Diffusion Implicit Models},
  author    = {Song, Jiaming and Meng, Chenlin and Ermon, Stefano},
  booktitle = {International Conference on Learning Representations (ICLR)},
  year      = {2021}
}

@inproceedings{lu2022dpmsolver,
  title     = {{DPM-Solver}: A Fast {ODE} Solver for Diffusion Probabilistic Model Sampling in Around 10 Steps},
  author    = {Lu, Cheng and Zhou, Yuhao and Bao, Fan and Chen, Jianfei and Li, Chongxuan and Zhu, Jun},
  booktitle = {Advances in Neural Information Processing Systems (NeurIPS)},
  year      = {2022}
}

@inproceedings{wu2023medsegdiff,
  title     = {{MedSegDiff}: Medical Image Segmentation with Diffusion Probabilistic Model},
  author    = {Wu, Junde and Fu, Rao and Fang, Huihui and Zhang, Yu and Yang, Yehui and Xiong, Haoyi and Liu, Huiying and Xu, Yanwu},
  booktitle = {Medical Imaging with Deep Learning (MIDL)},
  year      = {2023}
}

@article{wu2024medsegdiffv2,
  title={MedSegDiff-V2: Diffusion-Based Medical Image Segmentation with Transformer},
  author={Wu, Junde and Ji, Wei and Fu, Huazhu and Xu, Min and Jin, Yueming and Xu, Yanwu},
  journal={Proceedings of the AAAI Conference on Artificial Intelligence},
  volume={38},
  number={6},
  pages={6030--6038},
  year={2024},
  doi={10.1609/aaai.v38i6.28418}
}

@inproceedings{hatamizadeh2022swinunetr,
  title     = {{Swin UNETR}: Swin Transformers for Semantic Segmentation of Brain Tumors in {MRI} Images},
  author    = {Hatamizadeh, Ali and Nath, Vishwesh and Tang, Yucheng and Yang, Dong and Roth, Holger R. and Xu, Daguang},
  booktitle = {MICCAI Brainlesion Workshop (BrainLes)},
  year      = {2022}
}

@article{aboian2017imaging,
  title={Imaging characteristics of pediatric diffuse midline gliomas with histone H3 K27M mutation},
  author={Aboian, Mariam S. and Solomon, David A. and Felton, Eric and Mabray, Marc C. and Villanueva-Meyer, Javier E. and Mueller, Sabine and Cha, Soonmee},
  journal={American Journal of Neuroradiology},
  volume={38},
  number={4},
  pages={795--800},
  year={2017},
  publisher={American Society of Neuroradiology},
  doi={10.3174/ajnr.A5076}
}

@article{mi2025sfdiff,
  title={Diffusion network with spatial channel attention infusion for brain tumor segmentation},
  author={Mi, J. and others},
  journal={Biomedical Signal Processing and Control},
  year={2025}
}

@article{qin2025btsegdiff,
  title={BTSegDiff: Brain tumor segmentation based on multimodal MRI dynamically guided diffusion probability model},
  author={Qin, Jiacheng and Xu, Dan and Zhang, Hao and Xiong, Zhaoyu and Yuan, Yejing and He, Kangjian},
  journal={Computers in Biology and Medicine},
  volume={186},
  pages={109694},
  year={2025},
  doi={10.1016/j.compbiomed.2025.109694}
}

@article{wu2025fcfdiffnet,
  title={FCFDiff-Net: full-conditional feature diffusion embedded network for 3D brain tumor segmentation},
  author={Wu, Xiaosheng and Hou, Qingyi and Xu, Zhaozhao and Tang, Chaosheng and Wang, Shuihua and Sun, Junding and Zhang, Yudong},
  journal={Quantitative Imaging in Medicine and Surgery},
  volume={15},
  number={5},
  year={2025}
}
}

\end{document}